\documentclass[12pt]{spieman}  
\usepackage{amsmath,amsfonts,amssymb}
\usepackage{graphicx}
\usepackage{setspace}
\usepackage{tocloft}
\usepackage{url}

\usepackage{color,soul}

\title{Explainable AI for medical imaging: Explaining pneumothorax diagnoses with Bayesian Teaching}

\author[a*]{Tomas Folke}
\author[a]{Scott Cheng-Hsin Yang}
\author[a]{Sean Anderson}
\author[a]{Patrick Shafto}
\affil[a]{Rutgers University, School of Arts and Sciences Newark, Department of Mathematics and Computer Science, 101 Warren Street, Newark, New Jersey, 07102, USA}

\cftpagenumbersoff{figure}
\cftpagenumbersoff{table} 
\begin{document} 
\maketitle

\begin{abstract}
Limited expert time is a key bottleneck in medical imaging.
Due to advances in image classification, AI can now serve as decision-support for medical experts, with the potential for great gains in radiologist productivity and, by extension, public health. 
However, these gains are contingent on building and maintaining experts' trust in the AI agents. 
Explainable AI may build such trust by helping medical experts to understand the AI decision processes behind diagnostic judgements.
Here we introduce and evaluate explanations based on Bayesian Teaching, a formal account of explanation rooted in the cognitive science of human learning.
We find that medical experts exposed to explanations generated by Bayesian Teaching successfully predict the AI's diagnostic decisions and are more likely to certify the AI for cases when the AI is correct than when it is wrong, indicating appropriate trust.
These results show that Explainable AI can be used to support human-AI collaboration in medical imaging.

\end{abstract}


{\noindent \footnotesize\textbf{*}Tomas Folke,  \linkable{tomas.folke@rutgers.edu} }

\begin{spacing}{2}   

\section{Introduction}
AI has the potential to transform medicine, particularly in disciplines that are data-rich and rely on image-based diagnosis such as radiology \cite{jha2016adapting, yu2018artificial, pesapane2018_ai_radiology}.
One of the main constraints for such disciplines is specialist time. Modern image-classifiers could potentially help with the case load as they have comparable diagnostic performance to trained medical professionals \cite{liu2019comparison}.
If easy diagnostic decisions could be offloaded to artificial systems, this would allow human experts to focus on the most challenging cases, thus reducing costs, increasing treatment volume, and improving diagnostic performance.
However, because of issues relating to legal and ethical accountability, it is unlikely that artificial systems will make independent diagnoses in the immediate future \cite{pesapane2018_ai_reg}.
Instead, there will be human AI collaboration, where AI serve as decision-support system that helps human specialists by assisting with case-prioritisation or providing second opinions \cite{liew2018future}.

The efficiency of human-AI collaboration depends on clinicians understanding and appropriately trusting the AI system \cite{gunning2019darpa}.
However, the most accurate image classifiers, deep neural networks, are notoriously obscure, and hard even for AI experts to understand \cite{guidotti2018survey}.
Moreover, while accurate, image classifiers do make counter-intuitive errors and are susceptible to being fooled \cite{nguyen2015deep}. 
Thus, while their opacity is deeply intertwined with their success, it poses a challenge for applying deep neural networks on high stakes problems such as medical imaging---until we can develop methods that allow clinicians to develop understanding and appropriate trust. 

Explainable AI (XAI) addresses this problem by developing specialised \textit{explainer systems}, which have the aim of explaining AI inferences to human users \cite{arrieta2020explainable}.
However, most work in XAI has been done by computer scientists for computer scientists, resulting in technical explanations that might help engineers to develop better AI systems, but are of limited utility for other types of end-users \cite{miller2017explainable}. 
Indeed, of the many published approaches for explainable AI, a tiny minority have tested their efficacy in a user study \cite{lage2019evaluation}. 
The lack of systematic understanding of humans and AI as a system remains a challenge. 

Bayesian Teaching is an integrative framework and computational model that addresses these challenges. 
Rooted in the cognitive science of human learning \cite{Shafto2008,Shafto2014,Eaves2016-IDS}, and drawing upon deep connections to probablistic machine learning \cite{Murphy2012,Eaves2016-topic}, Bayesian Teaching authentically integrates models of human and machine learning in a single system. 
Bayesian Teaching casts the problem of XAI as a problem of teaching---selecting optimal examples to teach the human user what the AI system has inferred. 
Bayesian Teaching draws upon strong empirical foundations in cognitive science, where it has been proposed as a model to explain the speed and robustness of learning from others \cite{Eaves2016-IDS,Shafto2014}.

We will evaluate two forms of explanation that we have previously validated for explaining deep neural net image classification in ordinary (non-medical) images among non-expert participants \cite{yang2021mitigating}.
First, explanation-by-examples, where we select examples from the training data that are most likely to help the user predict the AI’s decision according to Bayesian Teaching.
Second, explanation-by-features, where saliency maps highlight the pixels in the image that most impacted the AI’s decision.
We combine these two methods so that users are not just exposed to influential examples, but also saliency maps that showcase which features the AI paid attention to in each example.
Our preliminary work showed that these two methods of explanation complement each other in that explanation-by-examples helps users identify trials when the AI was correct, whereas explanation-by-features improves error detection \cite{yang2021mitigating}, so we used both methods to optimize the explanations in this high impact use case.

To evaluate our system, we applied Bayesian Teaching to a deep neural net used to diagnose pneumothorax in x-ray images.
The explanations were integrated into an interface that carries basic functionalities for viewing x-ray images.
We designed an experiment that aimed to test (1) participants' understanding of the AI (captured by how well they could predict the AI's decisions) and (2) the development of appropriate trust (captured by when they chose to certify the of AI's decisions).
Radiologists were recruited to evaluate whether medical professionals benefited from the explanations generated by Bayesian Teaching.
Our results confirmed the utility of explanation both for understanding the AI and for developing appropriate trust in the AI system.

\section{Methods}

\subsection{Participants}
We recruited eight participants: One medical school student, one radiology resident from Mt Sinai Hospital in New York, and six radiology residents from Guy’s and St Thomas’s Hospital in London.
The radiology residents all had an average of three years of experience working as radiologists (Range: 1-10 years).

\subsection{Materials}
The experiment was conducted online so each participant completed the experiment on their own computer.
We maintained a standard image size to ensure comparability across sessions. To this end we enforced a minimum screen size of 1064 x 600 pixels.

\subsection{Bayesian Teaching}
Bayesian Teaching is a method for selecting optimal explanatory examples. Explanation-by-examples can be considered a social teaching act. The explainer (as the teacher) selects examples to guide the explainee's (as the learner) inference to a desired target.
In Bayesian Teaching a good explanation consists of a small subset of the training data that  will, with high probability, lead a learner model to the correct inference.
The general equation for Bayesian Teaching is 
\begin{align}
P_T(D|\Theta) 
= \frac{P_L(\Theta|D)P_T(D)}{\int_D P_L(\Theta|D')P_T(D)dD'}.
\label{eq:teaching}
\end{align}
In this equation, $D$ can be any subset of the training data; $\Theta$ denotes the inference of the AI to be explained; $P_T(D|\Theta)$ is the probability of choosing $D$ as the explanatory examples for the target $\Theta$; $P_L(\Theta|D)$ is the learner's posterior inference after receiving $D$; and $P_T(D)$ describes an \textit{a priori} bias for a certain kind of examples (e.g., favoring smaller subsets); and the integral is over all partitions of the training data consistent with $P_T(D)$.

In this paper, the target $\Theta$ is the target model's predicted label $c^*$ of a target image $\mathbf{x}^*$. The $P_T(D)$ spreads the probability uniformly over any $D$ that is composed of a true positive, a true negative, a false positive, and a false negative example from the training data, with labels predicted by the target model. The learner model, $P_L(\Theta|D)$, has exactly the same architecture as the target model, but whereas the target model is trained on the entire training data set, the learner model is only trained on the four examples in $D$. Given these specifications---the image-label pair $\{\mathbf{x}^*,c^*\}$, the constraint of $P_T(D)$, and a program that computes $P_L(\Theta|D)$---Equation~\ref{eq:teaching} outputs the posteriors on teaching sets $D$. A teaching set from the posterior is sampled by taking the maximum-a-posteriori $D$.

\begin{figure}[t]
  \centering
  \includegraphics[width=\columnwidth]{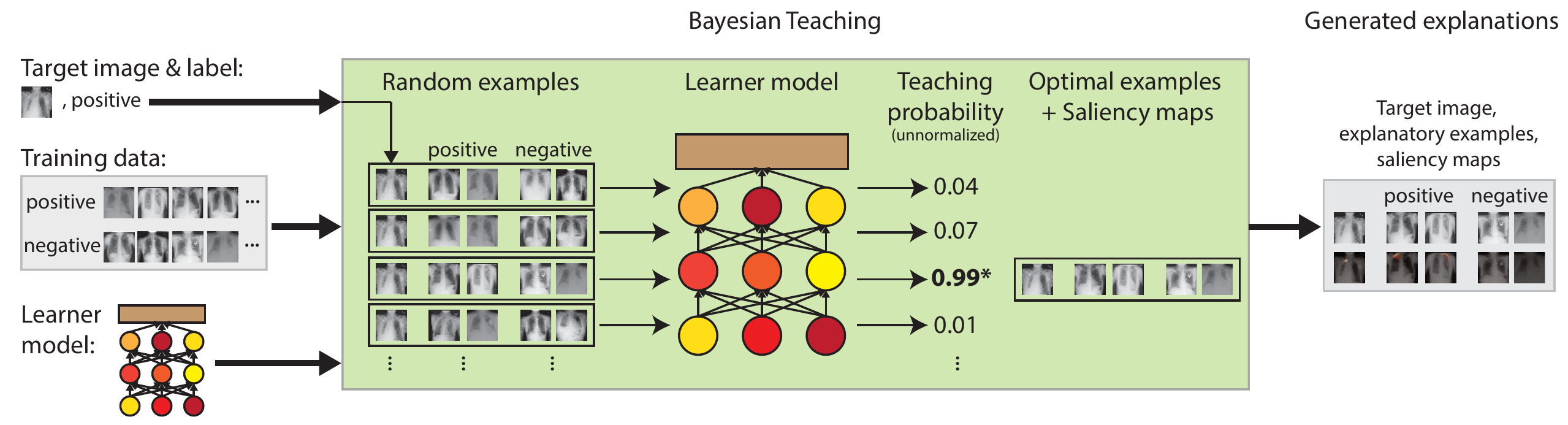}
  \caption{Selection of explanatory examples and saliency maps with Bayesian Teaching. The inputs to Bayesian Teaching are: the target image $\mathbf{x}^*$ and the label $c^*$ predicted by the target model; the training dataset; and the learner model, which is a program that computes $P_L(\Theta|D)$. The green box depicts the inner working of Bayesian Teaching. Image sets $D$ that satisfy $P_T(D)$ are sampled at random. The learner model takes in a large number of random trials (each row of $\{\mathbf{x}^*, D\}$) to produce the unnormalized teaching probabilities. Here, a set $D$ with probability $\approx 1$ is selected as the explanatory example. The final output is a set of ten images: the target image, the four examples selected, and the saliency maps of these five images. The saliency maps are generated from the target model (AlbuNet).}
  \label{fig:BT-flow}
\end{figure}

Figure~\ref{fig:BT-flow} shows a flowchart of how we implemented Bayesian Teaching. The inputs were the target image $\mathbf{x}^*$ and its label $c^*$; the training dataset; and the learner model, which is a program that computes $P_L(\Theta|D):=P_L(c^*|\mathbf{x}^*,D)$. Ten thousand sets of $D$ were sampled from $P_T(D)$. For each $D$, the (unnormalized) teaching probability, which is proportional to $P_L(\Theta|D)$, was computed. For $c^*=1$ and $c^*=0$, we randomly selected from sets of $D$ that satisfied $1-P_L(\Theta|D)<\epsilon$ and $P_L(\Theta|D)<\epsilon$, respectively, where $\epsilon$ is set to $10^{-6}$. 

The target mode, or the AI to be explained, is a deep neural network called AlbuNet used to diagnose pneumothorax in x-ray images \cite{Aimoldin2019}. AlbuNet was trained on x-ray images with radiologists' markings of regions of pneumothorax. The training data is the SIIM-ACR Pneumothorax Segmentation dataset hosted on Kaggle.\footnote{\url{https://www.kaggle.com/c/siim-acr-pneumothorax-segmentation}} To make inferences, AlbuNet  first computes the probability that pneumothorax is present for each pixel of the target image. It then takes these pixel-by-pixel probabilities (hereafter referred to as AlbuNet probabilities) and makes a binary classification for the full image by judging whether the number of pixels with AlbuNet probability greater than $b_1$ is greater than $b_2$.
We developed a probabilistic version of the original thresholding model. The new thresholding model has comparable accuracy (90\% vs the original 86\%) and is amenable to the probabilistic inference required by $P_L(\Theta|D)$ in Equation~\ref{eq:teaching}.


In contrast to the original model, the modified model applies soft thresholds in the form of logistic functions. Specifically, the probability that the image $\mathbf{x}$ has pneumothorax ($c=1$) under the modified model is given by
\begin{align}
P(c=1|\mathbf{x}, \theta) &= \lim_{\alpha\rightarrow\infty} \left(\prod_{j=1}^{M} p_j^{\alpha}
\right)^\frac{1}{\alpha} \label{eq:max_prob};
\\
p_j &= \frac{1}{1 + \exp^{-(w_1 x_{j1} - b_1)}} \cdot
\frac{1}{1 + \exp^{-(w_2 x_{j2} - b_2)}}.
\label{eq:2logits}
\end{align}
In Equation~\ref{eq:max_prob}, $\theta=\{w_1,b_1,w_2,b_2\}$ are the parameters of the two soft thresholds; $M$ is the number of pixels in image $\mathbf{x}$ whose AlbuNet probability are $>0.05$; $j$ indexes the pixels in image $\mathbf{x}$ that satisfy this criteria; and the infinity norm described by $\alpha$ picks out the pixels with maximal $p_j$ value in $\mathbf{x}$. In Equation~\ref{eq:2logits}, $x_{j1}$ denotes the AlbuNet probability for the $j^\mathrm{th}$ pixel in $\mathbf{x}$, and $x_{j2}$ is a normalized index between $[0,1]$ that measures that how many admitted pixels there are in $\mathbf{x}$ and how many of the admitted pixels in $\mathbf{x}$ are smaller than the $x_{j1}$. Note that $x_{j2}=1$ if all pixels in $\mathbf{x}$ are admitted and the corresponding $x_{j1}$ is the pixel with maximal AlbuNet probability in $\mathbf{x}$. 

To give an intuition for these equations, Equation~\ref{eq:max_prob} represents the image $\mathbf{x}$ by the pixels in it that have the highest probability for $c=1$. The product of the two logistic functions in Equation~\ref{eq:2logits} carves out a soft quadrant in the two dimensions represented by $x_{\cdot1}$ (the probability of pixel belonging to a region with pneumothorax according to AlbuNet) and $x_{\cdot2}$ (a normalized measure of the number of pixels with high AlbuNet probability) and assigns images with pixels in that quadrant a high probability to be $c=1$. Intuitively, an image will be represented in this quadrant of high probability for $c=1$ if it has a large number of pixels with high AlbuNet probability.

The $\theta=\{w_1,b_1,w_2,b_2\}$ of the target model was obtained by maximum likelihood training using gradient descent on the cross-entropy loss computed on the entire training dataset. The cross-entropy loss uses the $P(c=1|\mathbf{x},\theta)$ defined in Equations~\ref{eq:max_prob}--\ref{eq:2logits}. Similarly, the $\theta$ for the learner model is trained on the four examples in the selected $D$. The explanatory examples are selected when the $D$ produced a $\theta$ such that $P(c=1|\mathbf{x}^*,\theta)\approx1$ if $c^*=1$ and $P(c=1|\mathbf{x}^*,\theta)\approx0$ if $c^*=0$. The saliency maps of all images are simply the map of AlbuNet probabilities for that image. The saliency map is visualized using the hot colormap in Matplotlib \cite{Hunter:2007} with pure black and white fixed to AlbuNet probability $=0$ and $1$, respectively.




\subsection{Study Design}
\begin{figure}[ht!]
  \centering
  \includegraphics[width=\columnwidth]{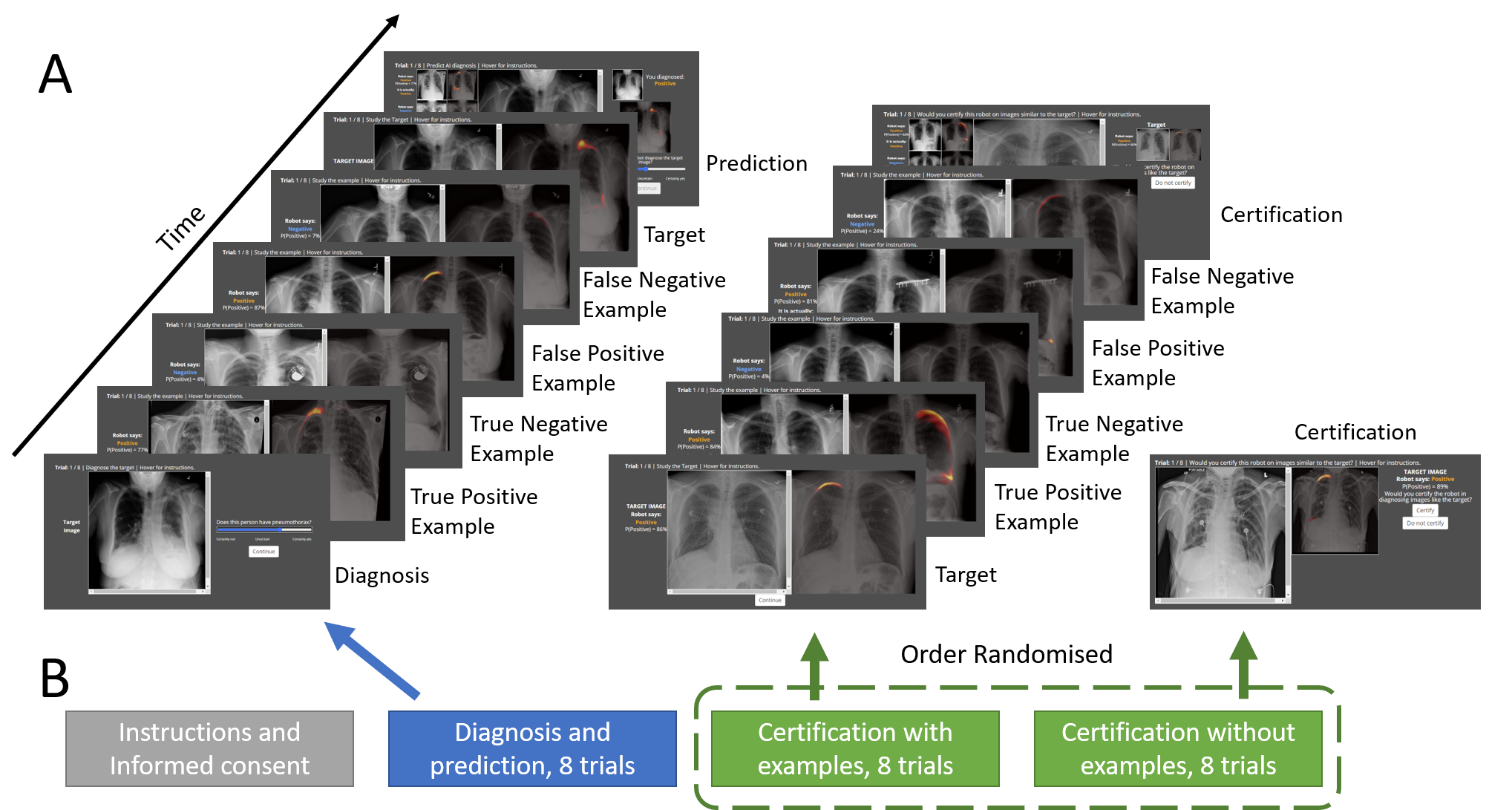}
  \caption{A. Trial structure for the different blocks. B. Experimental Design.} 
  \label{fig:procedure}
\end{figure}

Our experiment consisted of three blocks of trials, following consent forms and general instructions, see Fig.~\ref{fig:procedure}. The first block evaluated how well the participants could predict the AI diagnoses. The two subsequent blocks evaluated if the explanations developed appropriate trust by asking them to certify the AI for different cases. One of these blocks involved examples and saliency maps whereas the other just involved saliency maps.
Each block consisted of 8 trials, with target images counterbalanced based on the AI’s judgement, so that they included two each of true positives, true negatives, false positives and false negatives. In each block the presentation order of the trials was randomised and differed between participants.

Collectively the two certification blocks have sixteen target images, consisting of eight pairs. Each pair was constructed to belong to the same AI judgement category, and to be as similar as possible (in terms of L1 norm distance) within that category. The pairs were then split so that one member of each pair was shown in the block involving saliency maps but no examples, and the other was shown in the other block that involved both saliency maps and examples.  Which image in each pair belonged to each block was randomised and varied between participants.  

In the first block, each trial began with participants diagnosing the target image on a continuous rating scale with the endpoints labelled as “Certain pneumothorax present” and “Certain pneumothorax absent”. 
Responses were coded as an integer between 0 and 100, with 0 corresponding to certain absence of pneumothorax and 100 to certain presence.
They could zoom in on the target image and invert its colours, and they had unlimited time to make their judgement.
After making their diagnosis participants were shown four examples (one at a time). The examples were presented in a fixed order, starting with a true positive AI diagnosis followed by a true negative, a false positive and a false negative.
For each example participants were informed of the ground truth, the AI classification, and the AI probability judgement of pneumothorax.
Participants were also presented with a saliency map of the example image, and were able to zoom and invert colors of the example.
Participants decided how long to view each example.
Once participants had viewed all examples they were shown the target image with the AI's saliency map, after which we asked them to predict the AI diagnosis of the target image.
Participants made their prediction on a “reminder screen” that reminded them how they themselves had diagnosed the target, while also showing them miniatures of all the examples, and the target image, with the option to expand, zoom and invert any of the images. They made their prediction on a continuous rating scale similar to the diagnostic scale described earlier.
Once they had made their prediction they received feedback on the accuracy of their prediction.

The order of the two certification blocks were randomised between participants.
In one certification block participants were given the same information as in the prediction phase, in the other they only viewed a saliency map of the target image.
As opposed to the first block, in the certification blocks they were also shown the AI's judgement on the target image.
Participants were asked whether they would certify the AI for images similar to the target (a binary judgement), report whether they agreed with the AI's diagnosis, and finally justify their certification decision.
They could select multiple justifications among the following alternatives: (1) The robot got the correct answer, (2) The robot was appropriately confident, (3) The robot looked in the right place, (4) The examples are informative, (5) I am not certain I should certify, (6) Other. If participants selected (4)--(6) they had to elaborate in free text, but regardless of their choices they always had the option to elaborate in free text if they wished. 

\subsection{Statistical models}

\mathchardef\mhyphen="2D

All analysies were conducted in R (v 4.0.3).
Bayesian regression models were fitted in brms (v 2.14.0).
We used regularizing priors to ensure convergence: $\mathrm{Cauchy}(0, 10)$ for the intercept mean, $\mathrm{Normal}(0, 10)$ for the beta-coefficients, and
$\mathrm{Half}\mathit{\mhyphen}\mathrm{Student}\mathit{\mhyphen}\mathrm{t}(3, 0, 2.5)$ for the intercept standard deviation.
All models were sampled 2000 times in 4 different chains following 1000 warm-up samples per chain, resulting in 8000 posterior samples.
All posteriors were checked for chain convergence.

\section{Results}
\subsection{First order accuracy}
\begin{figure}[t]
  \centering
  \includegraphics[width=\columnwidth]{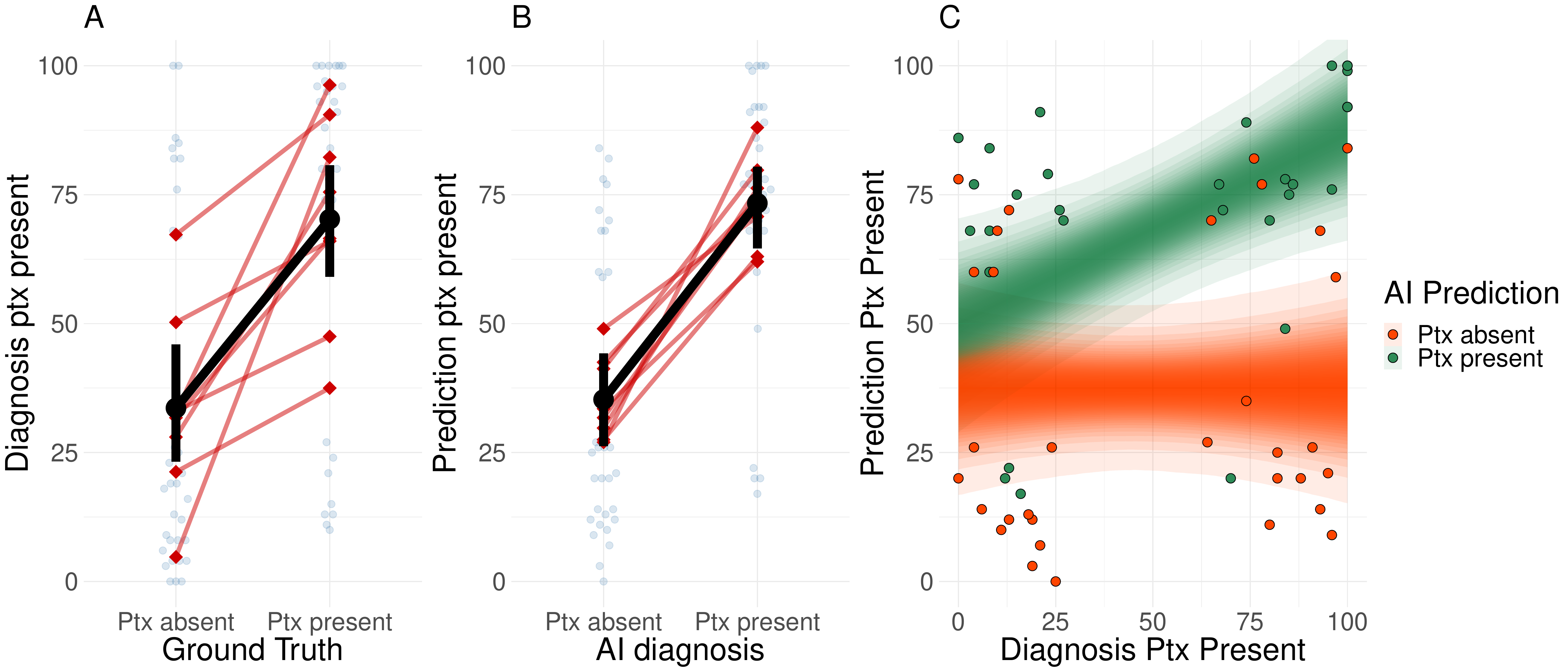}
  \caption{Diagnosing and predicting pneumothorax. A. The participants can reliably identify cases when pneumothorax is present. B. The participants can reliably predict AI classifications. C. The participants can reliably predict AI classification even when controlling for their own first order diagnoses. (A \& B) Light blue points capture unique trials; red diamonds and lines capture individual participants; and black lines and points capture the group level trends. Error bars represent 95\% bootstrapped confidence intervals. (C) Points indicate trial-level observations. The probability ribbon capture the posterior distributions of the regression lines from the hierarchical Bayesian model.} 
  \label{fig:diag_and_pred}
\end{figure}

To assess the first-order diagnostic accuracy of our radiologists we compared two nested Bayesian linear models: one predicting diagnoses from only participant-wise random intercepts (capturing participant-specific response biases) and one model that added a fixed effect for the ground truth of the target image (capturing the discriminant ability of the radiologists).
The second model fitted the data better than the first model, as indicated by a higher leave-one-out expected log point-wise predictive density ($ELPD_{loo}$) \cite{vehtari2017practical} than the first model ($\Delta$ = 6.5, se = 2.4), indicating that radiologists could successfully diagnose pneumothorax.
The posterior mean of this effect was 21.19 (95\% credible interval = 7.69--33.83) on a 100-point scale, suggesting that radiologist judgements (their diagnoses using the continuous rating) differed on average about 20 points between trials when pneumothorax was present and trials when pneumothorax was absent.
For a descriptive overview of these results see Fig.~\ref{fig:diag_and_pred} A.

\subsection{Predictive accuracy}
The radiologists correctly predicted the AI's judgement on 6 out of 8 trials on average (range = 5--7), see Fig.~\ref{fig:diag_and_pred} B.
Previous work on non-experts suggests that absent intervention humans expect the AI's judgement to mirror their own \cite{yang2021mitigating}.
To account for this we fitted three nested Bayesian linear models predicting radiologist predictions of the AI diagnoses.
The null-model contained participant-wise intercepts and a fixed effect of the radiologist's diagnosis for that trial.
The second model added a fixed effect for the AI's classification (coded as 0 for ``pneumothorax absent" and 1 for ``pneumothorax present").
The third model added an interaction term between the AI's classification and the radiologist diagnosis.
Radiologists could effectively predict the AI even when accounting for their first-order diagnosis as illustrated by the second model fitting the data better than the first model ($ELPD_{loo}$ $\Delta$ = 11.8, se = 3.5).
The third model fitted the data marginally better than the second model ($ELPD_{loo}$ $\Delta$ = 0.5, se = 2.3), implying that the first order diagnostic judgements might impact predictions of the AI differently when the AI is correct relative to when it is wrong, but we have too few observations to reach a strong conclusion.   

To more fully explore the relationship between radiologist predictions, radiologist diagnoses, and the classification of the AI, we studied the posterior coefficients of the third model, see also Fig.~\ref{fig:diag_and_pred} C.
Radiologists predictions tend to be more positive when the AI did classify pneumothorax than when it did not, even when they themselves found pneumothorax very unlikely (posterior mean = 12.47, 95\% credible interval = -2.38--27.22).
When the AI did not classify pneumothorax as present there was no relationship between radiologist diagnoses and their prediction of the AI (posterior mean = 0.00, 95\% credible interval = -0.22--0.22).
But for the trials when the AI classified pneumothorax as present there was a positive relationship between radiologist diagnoses and their prediction of the AI (posterior mean = 0.37, 95\% credible interval = 0.10--0.64).
The difference in intercepts indicate that the explanations worked: radiologist predictions were typically more positive on the rating scale for positive AI classifications than negative AI classifications, when accounting for their own diagnostic judgement. The difference in slopes indicate that the participants' own diagnoses serve as priors for the prediction of the AI for target images where the AI is correct, but not when it is wrong. 

\subsection{Certifications}
There are three key questions that we aim to address with regards to certifications:
1) Are participants more likely to certify the AI for images where it makes a correct diagnosis than where it makes mistakes?
2) Are they more likely to certify correct trials for the block with examples relative to the block without examples?
3) What justifications do participants provide for their certification judgements and what do these tell us about their decision processes? We address the first two questions with Bayesian regression models, to maintain analytic coherence. Because the third question is more qualitative and open-ended we only explore it descriptively.

To test whether radiologists are more likely to certify images that the AI classifies correctly, and whether the examples impact these judgements, we fit and evaluate three Bayesian hierarchical logistic regression models.
The null model predicted certification judgements (certified coded as 1, not certified coded as 0) from random intercepts at the participant level, the second model added AI correctness as a fixed-effect predictor, the third model added fixed effects for the explanation block and an interaction term between explanation block and AI correctness.

\begin{figure}[t]
  \centering
  \includegraphics[width=\columnwidth]{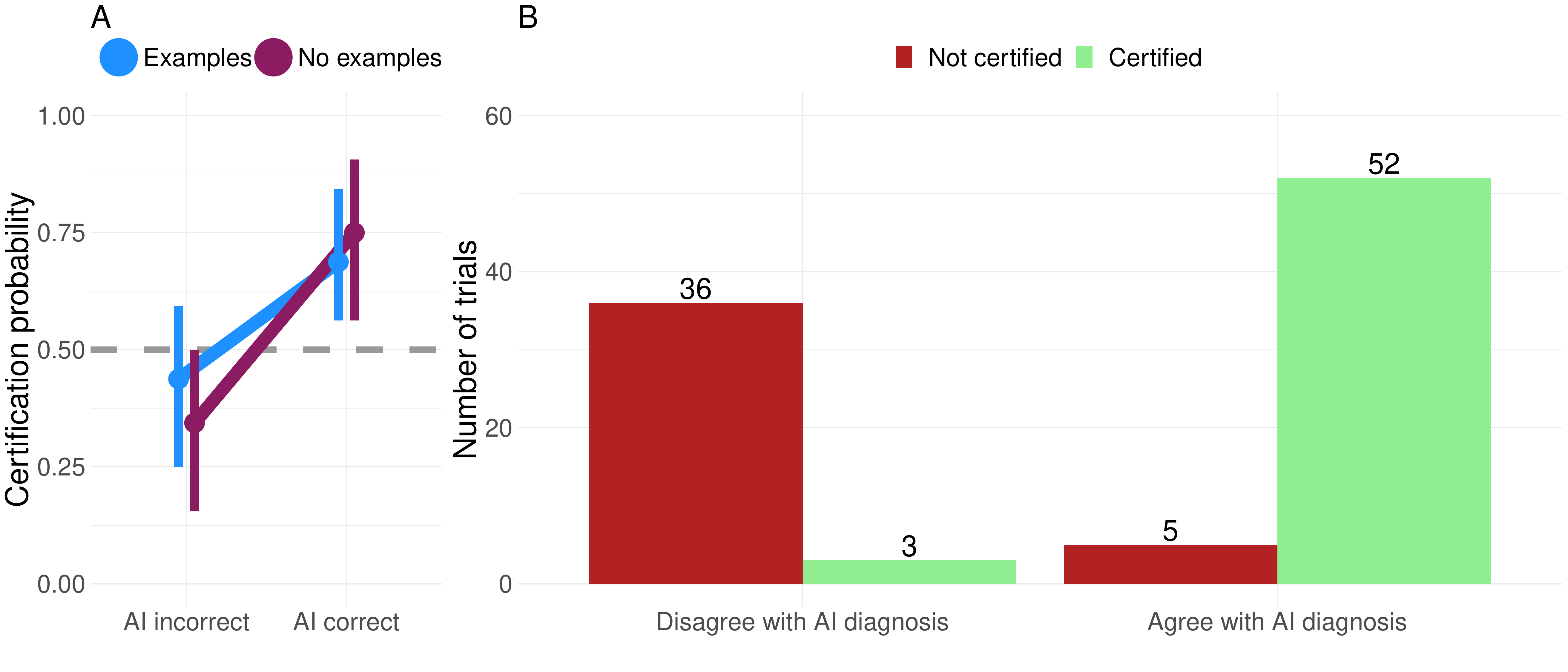}
  \caption{Certifying the AI. A. The participants are more likely to certify the AI for trials when it is correct than when it is incorrect, but the example block is not reliably different from the no-example block. The gray dashed line indicates chance performance, which in our task would be the result of always certifying the AI, never certifying the AI, or certifying at random. Error bars represent 95\% bootstrapped confidence intervals. B. Participants certification decisions are predominantly driven by whether they agree with the AI or not. Data are from 6 out of the 8 participants where we explicitly asked about their agreement with the AI's diagnosis.} 
  \label{fig:certification}
\end{figure}

The main-effect model accounted for the data better than the null-model ($ELPD_{loo}$ $\Delta$ = 7.0, se = 4.0) or the interaction model ($ELPD_{loo}$  $\Delta$ = 1.8, se = 1.1). These results imply that participants are more likely to certify trials when the AI classifies the target correctly (Mean OR = 5.12, 95 \% Credible interval = 2.21--10.72), but that there is no reliable difference in certifications between the block with examples and the block without examples, see Fig.~\ref{fig:certification}. Also note that certification probabilities tend to be below chance when the AI is incorrect but above chance when the AI is correct.
Because the two blocks are not reliably different we will collapse them in our subsequent discussion on certification justifications.

Our participants certified 46 cases where the AI was correct versus 25 cases when the AI was wrong.
They chose not to certify 39 cases when the AI was wrong, relative to 18 cases when the AI was correct.
In 16 out of 25 cases when participants certified the AI despite it being wrong, they justified their certification in terms of agreement with the AI. This suggests that these certification judgements are grounded in participant errors, which in turn implies that the ground truth is not a reliable proxy of participant belief in this task. Because we are interested in how radiologists justify certification in relation their own diagnostic judgements, we focus on 6 of the 8 participants that were explicitly asked if they agreed with the AI before they justified their certification decisions.

Certification decisions appear to be primarily driven by agreement with the AI,  see Fig.~\ref{fig:certification} B.
For the three cases where participants certified the AI despite disagreeing with it, they left open-ended responses clarifying their thinking. All of these responses suggest that they believed the AI actually got the overall classification correct, but had either been too confident or not confident enough regarding pneumothorax elsewhere in the lung, based on the saliency map.
In the five cases where participants chose not to certify, the most common justifications were either that the AI was looking in the wrong place or open-ended responses.
Here the responses again allude to getting judgements right for parts of the lung but making mistakes elsewhere, or that the AI has been performing poorly for a certain type of cases (e.g. lungs with prior surgical intervention or other pathology); therefore, despite the AI getting the particular case right, they would not want to certify it for similar images.
Collectively these results illustrate that the explanations enable radiologists to engage in complex reasoning about the AI judgements and capacity.

\section{Discussion}
In this study we have demonstrated that explanations generated by Bayesian Teaching help medical experts predict AI diagnoses and can be used to inform certification decisions.
There are two main limitations of this study: because of challenges associated with recruiting medical experts our sample size is modest, and we have only demonstrated the effectiveness of our framework for a single pathology (pneumothorax) and a single imaging type (x-rays). We are currently working to address both these limitations, as we are exploring additional recruitment channels to increase our sample for the current setup while designing variations of this experiment for other disorders (e.g. melanoma, breast cancer) and imaging types (e.g. CT scans, MRI, ultrasound).

Should our results generalise, they may have transformative impact on the medical domain by speeding up (and reducing the risks associated with) large scale uptake of AI as decision support in medical imaging.
The result would be reduced wait times for diagnostic tests and freed up expert time to focus on challenging cases.
Because Bayesian Teaching is a mathematical framework that is agnostic about which model is to be explained and the means of explanation, it can easily be adapted to different diagnoses, AI systems, and imaging types with little overhead.

The flexibility of the method also invites applications beyond medicine. This same method is applicable to other processes that require expert judgments about images including, for example, vehicular perception and self-driving in autonomous vehicles, insurance risk assessment, and cyber-physical systems in manufacturing \cite{varshney2017safety}.
Moving beyond image classification, Bayesian Teaching as a framework can improve AI support in any human endeavour including finance, logistics, and law. By demonstrating that Bayesian-Teaching-generated explanations can be used to teach domain experts to predict AI decisions and reason about AI judgements, we provide evidence of its suitability as an XAI method targeting end users in real-world settings. 

\section*{Disclosures}
The authors declare no financial interests linked to this manuscript and no other forms of competing interests.

\section*{Acknowledgments}
This material is based on research sponsored by the Air Force Research Laboratory and DARPA under agreement number FA8750-17-2-0146 to P.S. and S.C.-H.Y. The U.S. Government is authorized to reproduce and distribute reprints for Governmental purposes notwithstanding any copyright notation thereon.

This work was also supported by DoD grant 72531RTREP, NSF SMA-1640816, NSF MRI 1828528 to P.S.. The methods described herein are covered under Provisional Application No. 62/774,976.

\bibliographystyle{spiejour}   





\listoffigures

\end{spacing}
\end{document}